\documentclass[journal]{IEEEtai}

\usepackage[colorlinks,urlcolor=blue,linkcolor=blue,citecolor=blue]{hyperref}

\usepackage{color,array}
\usepackage{graphicx}
\usepackage[ruled,linesnumbered]{algorithm2e}

\begin{document}

\title{RoboSense At Edge: Detecting Slip, Crumple and Shape of the Object in Robotic Hand for Teleoprations} 
\markboth{Journal of IEEE Transactions on Artificial Intelligence, Vol. 00, No. 0, Month 2020}
{First A. Author \MakeLowercase{\textit{et al.}}: Bare Demo of IEEEtai.cls for IEEE Journals of IEEE Transactions on Artificial Intelligence}

\author{Sudev Kumar Padhi, Mohit Kumar, Debanka Giri and Sk. Subidh Ali \IEEEmembership{Member, IEEE} 
\thanks{This paragraph of the first footnote will contain the date you submitted your paper for review. This work was done on the dataset provided by Bharti School, IIT Delhi, under the 2022 Build-A-Thon ITU India AI/ML Challenge based on Low Latency Closed Loop. Our team secured 2nd position for the proposed solution in ITU India AI/ML Challenge and participated in the global ITU 2022 Build-A-Thon challenge. }
\thanks{The next few paragraphs should contain the authors' current affiliations, including their current address and e-mail. For example, F. A. The author is with the National Institute of Standards and Technology, Boulder, CO 80305 USA (e-mail: author@boulder.nist.gov).}
\thanks{S. B. Author, Jr., was with Rice University, Houston, TX 77005 USA. He is now with the Department of Physics, Colorado State University, Fort Collins, CO 80523 USA (e-mail: author@lamar.colostate.edu).}
\thanks{T. C. Author is with the Electrical Engineering Department, University of Colorado, Boulder, CO 80309 USA, on leave from the National Research Institute for Metals, Tsukuba, Japan (e-mail: author@nrim.go.jp).}
\thanks{This paragraph will include the Associate Editor who handled your paper.}}

\markboth{Journal of IEEE Transactions on Artificial Intelligence, Vol. 00, No. 0, Month 2020}
{First A. Author \MakeLowercase{\textit{et al.}}: Bare Demo of IEEEtai.cls for IEEE Journals of IEEE Transactions on Artificial Intelligence}

\maketitle

\begin{abstract}

Slip and crumple detection is essential for performing robust manipulation tasks with a robotic hand (RH) like remote surgery. It has been one of the challenging problems in the robotics manipulation community. In this work, we propose a technique based on machine learning (ML) based techniques to detect the slip, and crumple as well as the shape of an object that is currently held in the robotic hand. We proposed ML model will detect the slip, crumple, and shape using the force/torque exerted and the angular positions of the actuators present in the RH. The proposed model would be integrated into the loop of a robotic hand(RH) and haptic glove(HG). This would help us to reduce the latency in case of teleoperation.
\end{abstract}
\begin{IEEEImpStatement}
Today, haptic technology has emerged as a potential solution in robotics, like teleoperations, remote surgery, underwater exploration, manufacturing operations, remote gaming, Virtual reality and to name a few. Thus, it becomes essential to develop methodologies that can help us achieve high robustness and dexterity for the robotic hand in manipulation. This research work has been carried out with the aim of reducing the latency between the haptic glove and the robotic Hand. Many techniques proposed in the open literature mainly use multiple sensors to detect the slip of the object in the robotic arm. However, to the best of our knowledge, there is hardly any generalized (sensor-independent) techniques exist that can be easily integrated with different types of robotic hands. Also, integrating sensors like Bio-Tac or visual serving, IMU, camera, etc., increases the cost of deploying these robotic grippers on an industry level. Furthermore, the solution has to be lightweight for deploying at the MEC level. 
\end{IEEEImpStatement}

\begin{IEEEkeywords}
Slip detection, Deep Learning, Deep Neural Networks, Haptics, Robotic Hand, Long Short Term Memory, Teleoperation, 
\end{IEEEkeywords}


\section{Introduction}

\IEEEPARstart{S}{lip}, crumple detection and its prevention in robotic hands (RH) are essential for achieving the human level of dexterity and robustness in grasping. Humans rely entirely on tactile sensing feedback from the fingers to prevent an object from slipping or getting crumpled while grasping objects. The robotics community has always attempted to replicate the same response with a robotic hand. There is a plethora of research carried out to achieve the human equivalent dexterity in the robotic hand (RH). Slip detection has always been one of the most challenging research topics in the robotics manipulation community due to various issues, such as uncertainty in the hand's grip, and the latency in the loop of the haptic glove (HG) and the robotic hand (RH). Thus, achieving such a level of dexterity and robustness requires various parameters of the robotic hand to be considered, such as the number of actuators, the sensors present in the RH, the object's mass, its shape, and so on. Handling these parameters not only increases the complexity of the slip detection but also makes the slip detection problem challenging~\cite{1,3}.

In this work, we propose a machine learning (ML) based technique to detect the slip and crumple of the object present in the robotic hand. Adding to that, the shape of the object is also detected using our proposed technique.
We propose an intelligent slip-detection mechanism between the operator and the robotic hand based on machine learning by predicting the result of the commands being sent to the robotic hand during manipulation. The machine-learning model can predict the slipping and crumpling of the object in the robotic hand for every instance. Hence, the operator can use this feedback to change the force and angle accordingly to apply appropriate control and make the robotic hand grasp the object properly. 
This mechanism is implemented in the loop of the HG and RH, and it will help the HG operator control the RH efficiently and increase its dexterity. This feedback will also help the operator reduce the latency in the loop, which keeps the robotic arm from misinterpreting the commands, making it safe to perform critical tasks. This detection mechanism would be extremely useful in the field of teleoperated robotics,  where human error can occur possibly due to the delay in the commands. 
Similarly, it will also help an inexperienced operator to gain command over haptic gloves in less time and the chances of accidents will be significantly reduced.


Our proposed technique based on an ML model is generic sensor-independent and lightweight and can easily be integrated into scenarios where the robotic hand is being used for manipulation. 
However, integrating ML-based techniques should not increase the latency of the operations being performed by the robotic hand. Thus, the machine learning model should be lightweight (should require fewer features) and the inference time should be minimal. 
Our proposed ML model uses only the configuration of the actuators i.e. the time stamp data of force/torque and angular position of the robotic hand to detect the slip and crumple, which makes it computationally efficient by reducing inference time.
The target use case of our proposed technique is more beneficial when the robotic hand is used for performing tasks while connected to edge devices and operated remotely with the help of a haptic glove. The proposed technique can be deployed to the Edge devices, which will predict the slip and crumple during the manipulation of the object. This will monitor that the commands are sent and interpreted accurately in real-time without much delay to avoid any catastrophic incidents that happen due to the high latency between the HG and the RH.





\section{Background}
\subsection{Teleoperations}
Teleoperation refers to the context of operating machine like a robot from distance (remotely). Remotely operated robots can help us perform tasks that are not feasible for humans or sometimes not possible to attend in person since they pose some danger. There are certain environments that we can't directly manipulate because either they are located at a great distance or have contrived access as in the case of minimally invasive surgery~\cite{telerobotics}.. For example, in teleoperated surgery, a doctor in a different part of the world can perform a robotic surgery by operating the robotic hand remotely. During the COVID-19 outbreak, the use of teleoperated robotics increased significantly for contactless tests, medicine delivery, and disinfection~\cite{Covid-19}. 
Along the same line, in-person inspection or maintenance tasks are sometimes not feasible, like, in hazardous areas of deep mining fields or high-rise towers. In such scenarios, a remotely operated robotic hand is the only solution for any kind of manipulation.Remotely operated RH can also be helpful in space applications, where the astronauts would be required to manually perform the spacecraft's maintenance in zero gravity conditions. It can be a potential solution for remotely neutralizing explosive devices or handling nuclear wastes without risking human life~\cite{usecase}. In those cases, the RH can replace the need for human presence. 

\subsection{Haptic glove and Bilateral Teleoperation}
\begin{figure}[!htb]
    \includegraphics[scale=0.24]{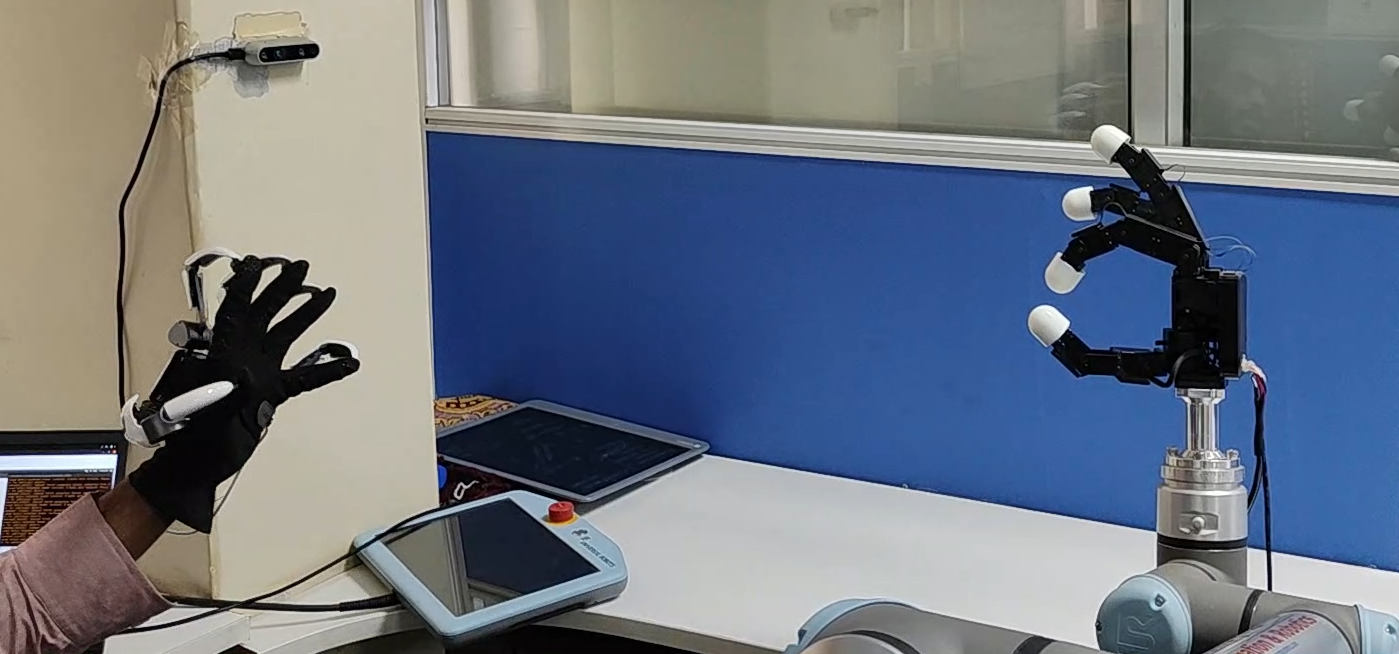}
    \caption{The operator is wearing the haptic glove and sending the 
    commands to the robotic hand}
    \label{fig:setup}
\end{figure}

With the rapid advancement in teleoperations with RH, the applications of HG have increased~\cite{haptics}. Haptics is one of the important aspects of tactile internet which helps the operator get a feel of touch and emulate the robot's behavior for the manipulation capabilities and gives a more immersive experience for the operator~\cite{13}. Bilateral teleoperation refers to the process of operating a machine at a distance by exchanging the action and reaction information between the master (the user or operator) and the slave (robotic hand) bidirectionally in real-time via a communication channel. In bilateral teleoperation, the user can give motion commands to the slave robot and can feel the interaction force through force feedback or haptic feedback ~\cite{Bilateral},~\cite{Bilateral2}. This allows more precise control to be able to restore the dextrous closed-loop manipulation. To perform bilateral teleoperation, a human operator wears the HG to operate the robotic hand remotely as shown in Fig-\ref{fig:setup}. The haptic glove uses the kinesthetic feedback of the human hand to predict the operator's intent, which is then transformed into the motion and movement (finger curvatures and joint angles for the robotic hand) of the robotic hand~\cite{13}. In this way, the operator sends the command using the haptic glove and replicates the manipulation behavior in the robotic hand, the operator can feel the force being applied on the object which makes a loop between the robotic arm and the haptic gloves. The operator acts as a human in the loop(HITL) and uses the feedback to operate the robotic hand.

\subsection{Teleoperation, latency and Multi-Edge computing,}
The current state-of-the-art teleoperation faces a major challenge in terms of latency in the loop of HG and RH. Traditionally, Teleoperation involves all the computation over the cloud which increases latency in the roundtrip of command and feedback. This latency can sometimes lead to mishaps, especially in critical applications like remote surgery. To tackle this limitation, the use of Edge computing for data processing has been in development. With the rise of demand for low latency loops, edge computing became the de-facto standard of the industry and shifted focus towards Egde AI/ML. Edge computing devices help to move our services from the cloud server to the edge~\cite{mec}. Multi-edge computing (MEC) makes the service closer to the customer by minimizing the amount of long-distance communication. It helps to reduce the latency in the loop and bandwidth use since the collection and processing of data become faster~\cite{mec2}. The ML model will be deployed directly to the edge device to reduce the latency. However, edge computing has its own limitations in terms of computation, bandwidth, Space and Power Constraints, and  Support Integration with Networks due to limited resources and investment~\cite{mec3},~\cite{mec4}. Thus, there is a need for edge computing that lightweight technique requiring less computation has to be developed, and generally, ML models are heavy in computation. Keeping this limitation in mind, we have designed the ML model for slip and crumple detection, which can be deployed on edge devices level so that latency that arises during computation on the cloud server can be reduced as well as the required operation can be performed efficiently.

\subsection{Challenges in Teleoperation with Robotic hand}
In this section, we highlight some other key challenges in teleoperation with RH and HG which would also be addressed by this proposed technique of feedback mechanism. 




\begin{itemize}
    \item \textbf{Lack of feedback:} One of the major challenges is a lack of feedback or limited feedback in the loop of RH and HG. Humans can easily perform very complex, dexterous manipulation tasks by reacting to tactile sensing observations. Humans form a force loop through the muscles and a position loop through the eyes. In contrast, robots cannot perform reactive manipulation due to a lack of sensor information and mostly operate in an open loop while interacting with their environment. The current state of the art of telerobotics involves supervisory control by visual data obtained from the camera. Since the camera transfers the visual feed with a delay of about $300$-$400$ms. This makes the command to be sent sequentially i.e. move and wait hence making the whole process quasi-static. Also, the user's intent prediction has to be done before sending it to the robotic hand. Processing this data and transferring them over long distances via cloud services increases the latency in the loop. There are other feedback systems proposed in the literature based on force feedback or tactile feedback~\cite{feedback2}. However, they have their own limitations in their stability and transparency~\cite{feedback}. In order to execute tasks remotely that require human-level precision, it is essential to make both haptics and visual feedback available to the operator. Consequently, the current state-of-the-art manipulation algorithms are not that efficient in performance and can only work in highly structured environments with the help of sensory feedback.  
    
    \item \textbf{Lack of Dexterity: } A major challenge faced in performing tasks using a robotic hand is achieving dexterity close to that of a human hand. The human hand uses the whole hand surface for grasp, while the robotic hand uses only the fingertips thus, the robotic hand has limited capabilities for performing manipulations. For example, while grasping the object if the robotic hand cannot apply the right amount of force, the object may slip out of hand or get crumpled leading to some catastrophic incident. Along the same line, While performing a medical surgery remotely, if the robotic hand performs some operations in the wrong sequence or loses control over any surgical equipment, it can put the patient's life at stake~\cite{telesurgery}. 
    
    \item \textbf{Skills of the haptic operator}: Another challenge is the difference in operator's levels of experience over the haptic glove and it is time-consuming to learn and master by an inexperienced.  For a novice operator, estimating the required force to be applied to the object in hand without slipping or crumpling is hard. There are two main reasons behind this; the first one is due to the differences in the degree of freedom of the robotic hand and the haptic glove, and the second one is the inertia of an object varies with its shape and size~\cite{0}. This makes it difficult to gain command over haptic gloves to operate a robotic hand.
    
    \item \textbf{Slip and Crumple detection:} While grasping an object or an instrument, it becomes essential to detect the slip and crumpling of the object present in the RH so that the object can be firmly held to perform the desired task. This proposed intelligent feedback will help the operator estimate the optimal force required for holding an object to perform an operation. Furthermore, Detecting object shapes is extremely useful for the operator to control the gripper much more efficiently. If the model can detect the object's shape in the robotic hand, the haptic glove input can be modified accordingly for robust grasping of the object. This will help the operator to manipulate the object without the need for a camera feed or in an unknown environment. In this work, we trained our Machine learning model for general shapes like spheres, cubes, cuboids, or rugby with different dimensions. However, training the model can extend the technique to other complex shapes. 
\end{itemize}

\subsection{Machine Learning and Slip Detection}
Machine learning-based techniques for slip detection have been in development for a long time. Analytical modeling gives satisfactory results in slip prediction, but modeling a multi-fingered gripper becomes difficult, also it cannot cover all of the edge cases. Hence, the machine learning-based model for slip detection is the choice to cover all the parameters. With the usage of various sensors like IMU, Tactile sensor, and camera, the data is gathered to train the ML model.  
In order to mimic human-like skin tactile-based sensor is used, and a support vector machine (SVM) is used to detect to identify slip of the object from the gripper ~\cite{1,3}. 
However, as there are multiple fingers, the prediction for slip detection becomes very complex, and it does not yield very good accuracy. Furthermore, it also increases the latency in the loop. 
Interpreting tactile information using an Optical-tactile sensor (TacTip) on the fingers of the robotic hand is also being used~\cite{optical}. This technique uses ML models, such as  SVM and logistic regression, to detect the slip of the object from the $3$-fingered UR$5$ robotic arms. However, the feedback loop is considered to be relatively very slow in this case. 
Bio-Mimitic sensors are also used to get the velocity of the internal pin of the Tic-tac sensor, this data is used to train SVM to predict the slip~\cite{3}. Furthermore, accuracy is improved by utilizing the accelerometer and piezoresistive pressure sensors to measure grasping force~\cite{4}. 
Techniques are also developed to detect the slip direction in a gripper 
Using a bio-tactile sensor, the data is obtained from the sensor to train a Deep Neural Network (DNN) to predict the slip direction and prevent it by applying the appropriate action~\cite{2}. 
Moreover, a precise accelerometer sensor is employed to find the acceleration signal of the slipping object in the presence of gravity.

Later techniques use the information from the Gelsight tactile sensor~\cite{5}, a camera-based tactile sensor, with an external camera. The data obtained are used to train DNN to detect the possibility of slip. The combined data provides better accuracy than any of the individual sensors.


The ML-based techniques rely on some sensor feedback and training of some ML models to process the data for accurate slip prediction. Using the SVM and training data from the sensors fails to achieve high accuracy. With the introduction of DNN, the accuracy increased to a certain extent. However, since the data from the sensors are time series data, the object may be held properly at an instant and may slip at the next instant, which means Slip occurs for an instant only. Hence, the DNN model fails to generalize that due to a lack of extraction of the important features that are responsible for slip detection. 

Moreover, the models mentioned above need information for identifying slip detection and very specific sensors.

Our proposed technique will solve the problem of lack of tactile feedback, which includes slip and crumple detection, and identify the object's shape accurately without using tactile sensor data.  Our proposed technique requires only available information about the object and the configuration of the actuators to predict the object's shape correctly and employ slip and crumple detection. Using this technique, we can identify important features such as slip and crumple,  which will make the  Loop between the haptic glove and the allegro hand more stable and robust by controlling delays and misinterpretation. Unlike other mentioned techniques, we have designed our model to work in real-world scenarios such that the robotic arm can be made more reliable with the help of machine learning.

The scope and advantages of our proposed technique can be summarised as follows:
\begin{itemize}
   \item Machine learning model will act as the feedback by detecting the important features of
   grasping like slip, crumpling, and shape, this will help the operator to modify grasp.
   \item Proposed technique will help increase the dexterity of the robotic hand by making the control more robust. 
  \item  The  ML model in our technique uses only the actuator's data which makes it cost-effective and computationally ideal for MEC-based teleoperation applications.
    \item  It will minimize the error that arises due to novice HG operators and different configurations of HG.
\end{itemize}

\begin{figure}[!htb]
    \includegraphics[width=0.50\textwidth]{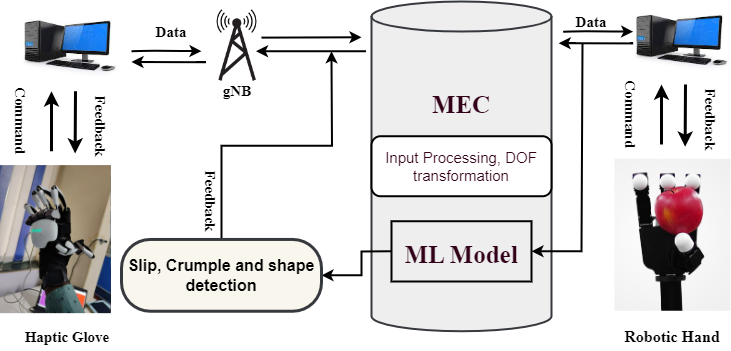}
    \caption{The Architecture of the Robotic Hand and Haptic Glove}
    \label{fig:MEC}
\end{figure}



 \section{Objective}
We have divided our technique into two sub-problems for creating a low latency closed loop between the haptic glove, ML Model, and the robotic hand (Allegro Hand), which will be deployed in the MEC bed to minimize the latency. The two sub-problems which solve the issues mentioned are:\\
\begin{enumerate}

    \item \textbf{ Slip and crumple Detection}:\\
    Given an object presents an allegro hand, the object may tend to slip( Fig-\ref{fig:slipping} and Fig-\ref{fig:non_slip}) or crumple from the grip. Using the machine learning model, our goal is to predict if the object is slipping out of robotic hands or getting crumpled. We use the time-series data of the Allegro hand, which includes the force/torque and angle made by the Allegro hand's sixteen actuators along with the mass and shape of the object in hand.  This data is used to predict the values of slip and crumple as a multi-output prediction. The slip's value can be either zero ( not slipping) or one (slipping). Similarly, the crumple's value can be either zero ( not crumbling) or one (crumbling). Therefore the multi-output can have four different combinations. This output has to be provided for every instance of the given data.

\begin{figure}[!htb]
    \centering
    \includegraphics[width=0.48\textwidth]{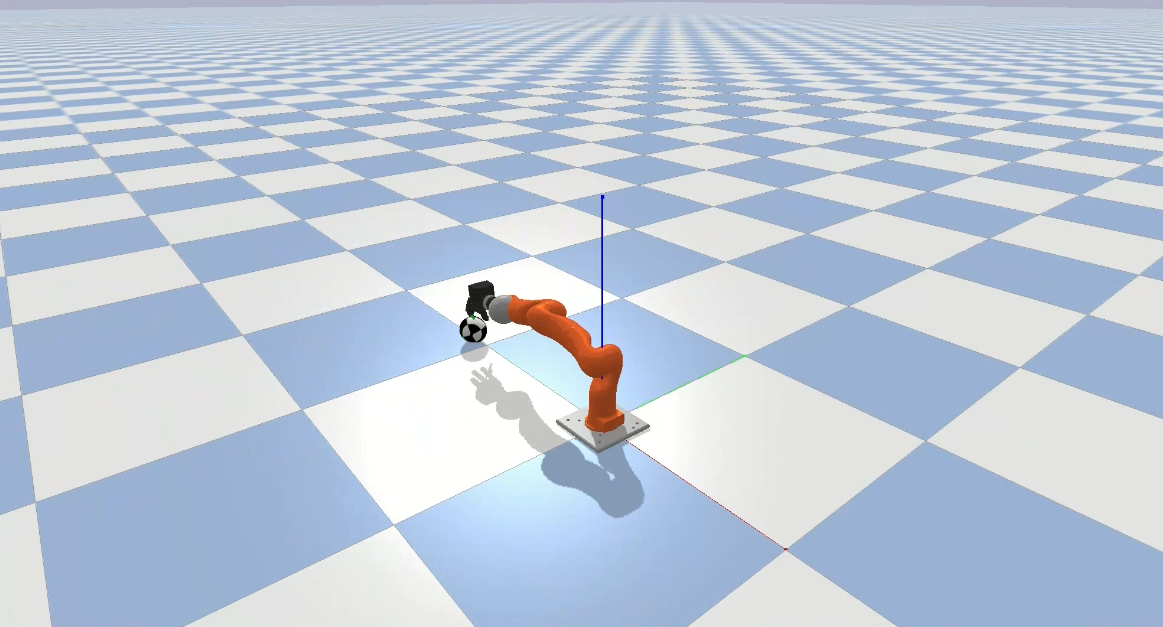}
    \caption{while picking the object, it slipped due to less force }
    \label{fig:slipping}
\end{figure}

\begin{figure}[!htb]
    \centering
    \includegraphics[width=0.48\textwidth]{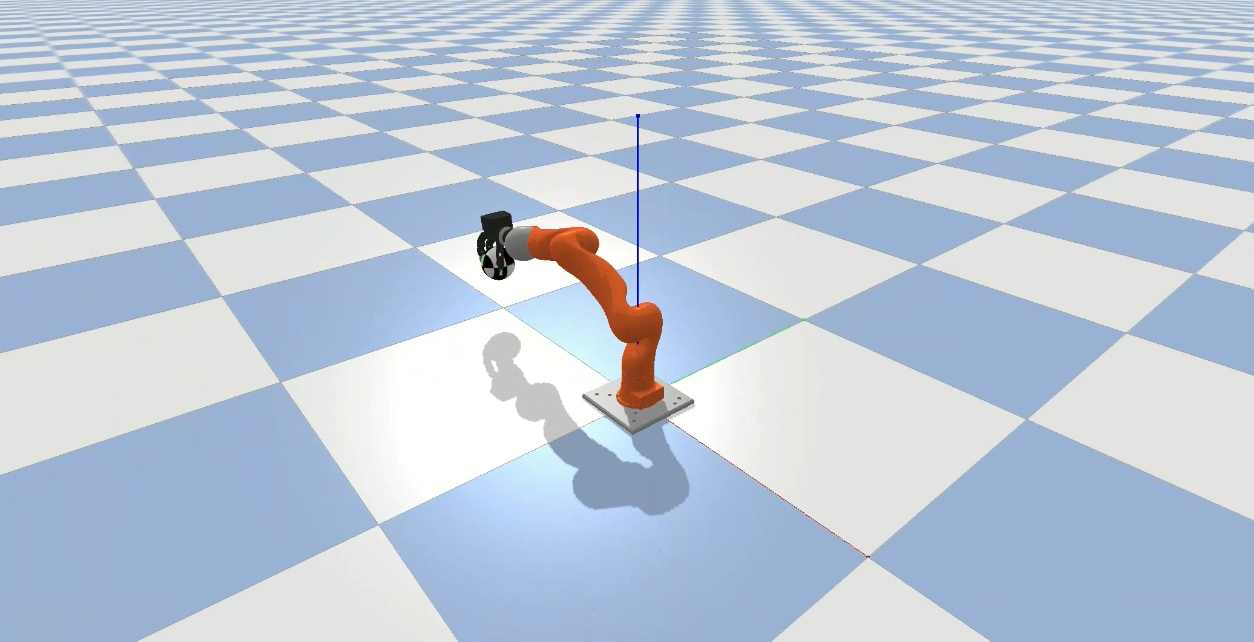}
    \caption{Object picking without any slip or crumple}
    \label{fig:non_slip}
\end{figure}

 \item \textbf{ Object Shape Detection}:\\
 The object is currently present in RH, our aim is to predict the shape of the object from $13$ generic shapes using the Machine learning model. The ML model is given the force of each actuator and the angle that the object is making with the $16$ actuators of the Allegro hand, as well as the mass of the object as input. The model will predict one of the $13$ object shapes as the output. The number of objects can be increased based on the task and requirements. 
 
\end{enumerate}


\section{Methodology}


The proposed technique comprises training an ML model to predict the slip and crumple as well as the shape of the object present in the RH. The training dataset is timestamp data obtained from the actuators of the RH, which is controlled using an HG in the MEC setup. As already mentioned, our proposed techniques focus on two aspects of the RH where the operator controls it, i.e. \textbf{a)} slip and crumple detection and \textbf{b)} object shape detection. Now, we will discuss our approach for each of the mentioned aspects.


\subsection{Slip and Crumple Detection}

When the operator controls an RH, there might be an instance where the force applied to the object present in the RH is less or more such that the object is slipping or crumpling from the grip. This can occur due to a variety of reasons, such as the operator's adaptability to different haptic gloves, lack of quick feedback due to the abundance use of sensors, etc. Our proposed technique will be detecting the slip and crumple of the object present in the robotic hand using a lightweight ML model and the quick feedback can be used by the operator to take control of the circumstance. The ML model only uses the data from actuators along with the mass and the shape of the object to predict the state of slip and crumple correctly. This makes our technique fast, efficient, and lightweight, which can be deployed in edge devices. 

We used a Multitask Classifier based on the encoder architecture of Tabnet~\cite{tabnet} to classify slip and crumple. This model encodes the features with the help of a feature transformer, an attentive transformer, and feature masking. These features are passed into a fully connected layer to find the best and the most dominant set of features to detect the slip and crumple. Here, we use multiple decision blocks, each of which focuses on processing a different subset of the input features.  The feature transformer unit and the attentive transformer unit obtain an encoded feature as output. Now, the encoded feature is fed to the feature selection mask unit for selecting the features of the object for acquainting purposes. The selected features of the object for acquainting purposes are fed again to the feature transformer unit to provide another output. Now, the processed output is divided by a split block unit for being used so that one part is used by said attentive transformer unit in the next step and the other part to get a final output. The final output is fed to a rectified linear activation function to get another output. Now, all the beforehand processes are performed three times for aggregating all the finally obtained outputs, and all the aggregated output is fed to a fully connected layer for processing and acquiring said robotic hand (2) with all slipping and crumpling possibilities and all shapes of said object that is used while teleoperated functions are being carried out remotely.


We have tried out simple DNN and LSTM-based models to classify slip and crumple. The simple DNN failed as the slip and crumple for a particular object occurred continuously, which the simple DNN model did not capture. Then we moved on to the LSTM-based model as the data was continuous.


However, LSTM failed as the slip or crumple was occurring for a very small time frame,  and then the state of slip and crumple changed in an instant, so building the relationships between slipping and crumpling of different types of objects was hard to form. Then we chose a model that was able to solve the problem by finding more dominant features, which gave us a higher accuracy.

\begin{figure}[!htb]
    \centering
    \includegraphics[scale=0.35]{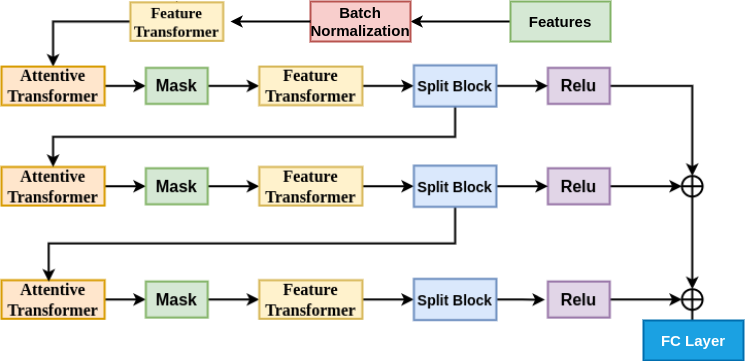}
    \caption{Caption}
    \label{fig:my_label0}
\end{figure}

Our model consists of the following modules:
\begin{itemize}
    \item Feature transformer is composed of a fully connected (FC) layer, BN (Batch Normalization), and GLU (Gated Linear Unit) nonlinearity which converts features into more interpretable attributes. 
    \item Attentive transformer blocks consist of a single layer mapping modulated with prior scale information, aggregating how much each feature has been used before the current decision step. Here, the coefficients are normalized using sparsemax, leading to a sparse selection of prominent features.
    \item Feature selection mask gives interpretable information about the operation of the model, and the masks can be used to get the required global feature attribution.
\end{itemize}

\begin{figure}[!htb]
    \centering
    \includegraphics[scale=0.6]{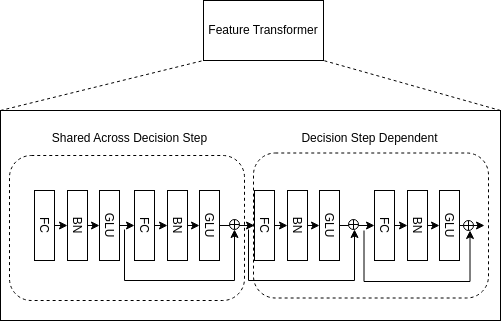}
    \caption{Caption}
    \label{fig:my_label1}
\end{figure}

\begin{figure}[!htb]
    \centering
    \includegraphics[scale=0.6]{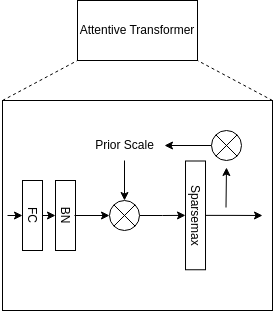}
    \caption{Caption}
    \label{fig:my_label2}
\end{figure}

Input features are passed through the feature transformer and attentive transformer before finding the feature selection mask. Then, on finding the mask, the encoded features are again passed on to the feature transformer. A split block divides the processed representation to be used by the attentive transformer of the subsequent step as well as for the overall output, which is passed onto the RELU function. This is performed three times, and all the outputs of these layers are added and passed on to a fully connected layer. 
Note: We have also used simple deep neural networks, LSTM, to train, but the accuracy of the proposed technique was highest due to the reason that it encodes the features into a more meaningful entity, which correlates to the output labels more.

\subsection{Shape Detection}
Given the object is in grasp the ML model has been trained to detect the shape. For shape detection, the dataset includes the position and force of 16 actuators of the hand, and the weight of the object. This data has been used to train a model to detect the shape of the object. Following are the steps involved in Feature Engineering:
\begin{itemize}
\item  First, we used a label encoder to encode the column Object Held, which consists of the shape of the objects. It contains alphabets, so we changed it to labels 0 to 12 for the 13 objects' shapes.
\item  Then we split the data as X and Y, where X is the data that will be given to the model, and Y is the actual output label of the object shape.
\item  Then, we scale the data with the min-max scaler. 
\item  Then we shuffle the data and split it before giving the data to the model.

\end{itemize}

The model we used is a simple Deep Neural Network that consists of 5 fully connected (FC) layers. We have used 10-fold cross-validation to increase the accuracy and get the best parameters. We tried out other models but didn't get any drastic improvement, and this small model was giving similar results, so we went with the smaller model. We tweaked the number of nodes in the layers to choose the final model and got the best result.

\section{Experimental setup}

\subsection{Dataset}
\begin{figure}[!htb]
    \centering
    \includegraphics[scale=0.06]{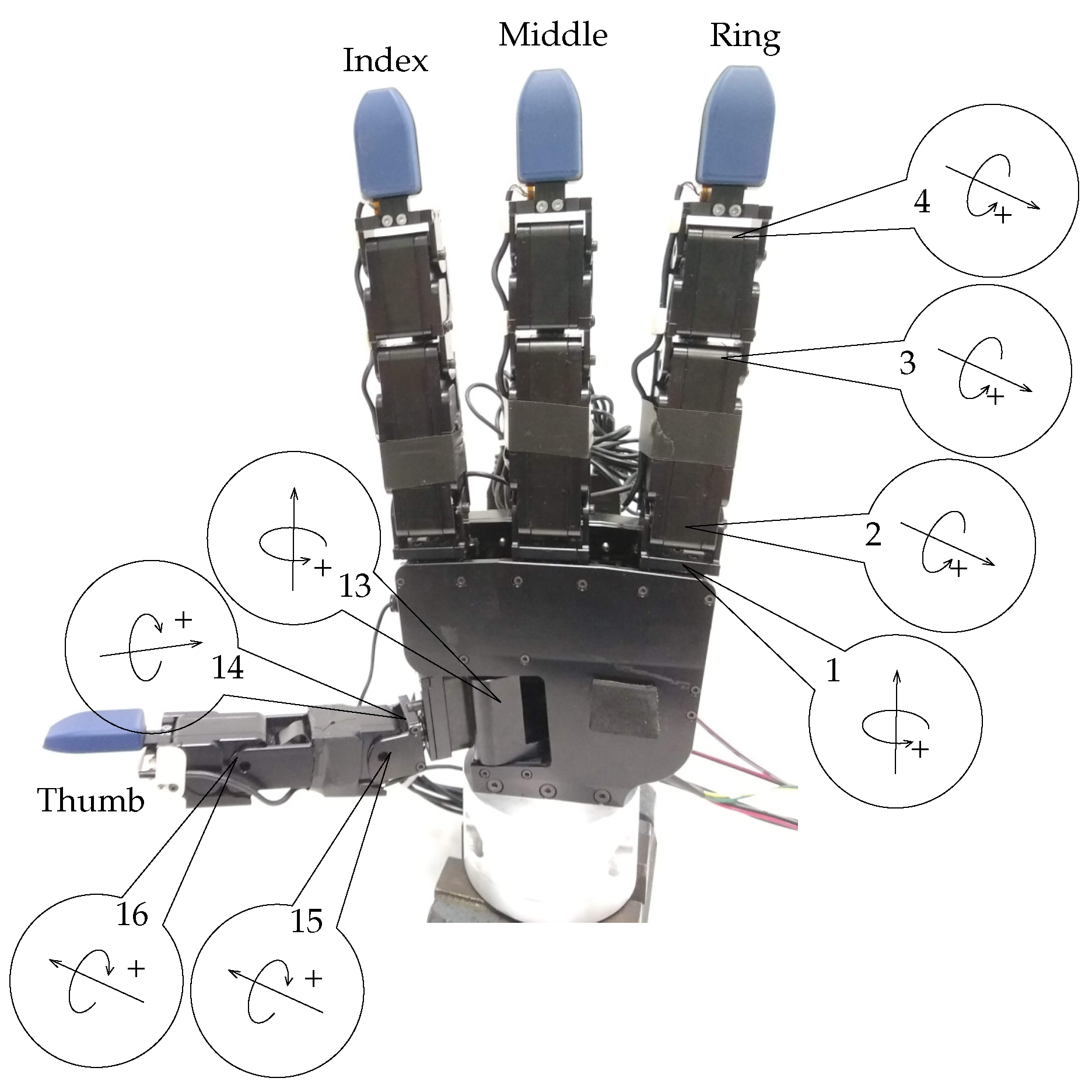}
    \caption{Allegro Hand discription}
    \label{fig:AH}
\end{figure}

The data used to train our ML model consists of timestamp data of the angle and force of $16$ actuators in the robotic hand, along with the mass and shape of the object. The output label of the dataset is the status of slip and crumple, i.e. $o$ for false and $1$ for true. Thus, there are four combinations possible. 
For offline training of the ML model, the data has been generated using the Pybullet simulation which is a physics-based engine. The simulation environment used an Allegro hand gripper, which consists of three fingers and one thumb, $16$ actuators in all ($4$ in each finger)Fig-\ref{fig:AH}, and the KUKA IIWA robotic arm. When an operator is doing some manipulation with the HG, which controls the allegro hand, we get the force and angle measurement along with other parameters that depend on the object, like the shape and Mass, which has been used to train the machine learning model. Several types of objects, like spheres, cubes, and rugby balls, with different dimensions, have been used to generate the data. The training data consisted of the joint configuration of the 16 actuators at different time stamps, i.e., force/torque applied at each joint and their angular positions with the object held. We have used the same simulation environment for simulation, where we provided the torque and angular positions to each joint through a Python script and simulated its grasping. The output of the machine learning model is consistent with the output of the simulation environment. The proposed technique would be integrated with RH which is connected with MEC, ML model would be deployed at MEC in the loop of RH and HG to predict the slip, crumple, and shape of the object(Fig-\ref{fig:MEC}).
 
 
The following are the steps we followed for feature engineering:
\begin{itemize}
%
    \item Use a label encoder to encode the column named {\em size} as it contains alphabets, so we changed it to labels $0$, $1$ and $2$ for three object sizes, i.e., $5\times10\times5$, $R3.5$ and  $5\times5\times5$.
    \item Removed the timestamp column as we tried our techniques both with and without the timestamp, and the latter technique was more accurate. 
    Note: The dataset used has different timestamps for each event, making the impact of timestamps less related to the output (slip or crumple).
    \item Divide the data into $X$ and $Y$, where $X$ is the data given to the model, and $Y$ is the actual output label for slip and crumple.
    \item Scale the data with the min-max scaler. 
    \item Split the data into a train and test set by shuffling the data and giving the data to the model
\end{itemize}

\section{Results And Discussion}
The below-shown tables are the accuracy results of the model training. 
\begin{table}[!htb]
\centering
\begin{tabular}{|c|c|c|}
\hline
\textbf{Models}              & \textbf{\begin{tabular}[c]{@{}c@{}}Slip Detection   Model\end{tabular}} & \textbf{\begin{tabular}[c]{@{}c@{}}Shape Detection  Model\end{tabular}} \\ \hline
\textbf{Validation accuracy} & \textbf{93.2}                                                             & \textbf{99.7}                                                             \\ \hline
\textbf{Training Accuracy}   & \textbf{93.7}                                                             & \textbf{99.8}                                                             \\ \hline
\textbf{Loss}                & \textbf{0.13521}                                                          & \textbf{0.014}                                                            \\ \hline
\end{tabular}
\end{table}

\begin{table}[!htb]
\begin{tabular}{|c|c|c|}
\hline
\textbf{Hyperparameters}  & \textbf{Slip Detection Model} & \textbf{Shape Detection Model} \\ \hline
\textbf{Framework used}   & Pytorch                       & Pytorch                        \\ \hline
\textbf{Learning rate}    & 0.02                          & 0.002                          \\ \hline
\textbf{Optimizer}        & Adam                          & Adam                           \\ \hline
\textbf{Loss Function}    & CrossEntropy                  & CrossEntropy                   \\ \hline
\textbf{Model Type}       & Deep Neural Network           & Deep Neural Network            \\ \hline
\textbf{Epochs}           & 100                           & 200                            \\ \hline
\textbf{Validation Split} & 0.16 percent of whole data    & 10 Fold validation             \\ \hline
\textbf{Batch Size}       & 2048                          & 2048                           \\ \hline
\end{tabular}
\end{table}

The results make it clear that the shape detection algorithm has higher accuracy than the slip detection since the shape of the object is very particular, which gives us a well-defined configuration of the actuators for a particular shape. In the case of slip detection, the machine learning model has an accuracy of $93$ percent, which can be improved by training the model for more data to cover some edge cases as well. Also, this technique can be extended for other shapes by training the model for different shapes of objects.

\section{Conclusion and Future works}
The presented technique is an implementation of a Machine learning model in the loop of RH and HG at the EDGE level, to detect the slip, crumple, and shape of the object. The prediction of the Machine Learning model will help the HG operator to modify the grasping strategy for a dextrous manipulation of the object. We use Pybullet for the simulation in real-time to demonstrate the slip and crumple. We intend to extend the implementation of this technique on real hardware, i.e., KUKA Arm and Allegro which are integrated with MEC as future work. Moreover, the Machine learning model would be trained better by generating more data for different types of shapes and sizes of the objects so that this implementation can be extended to industry-level usage.  

\section*{Acknowledgment}

We would like to thank IIT DELHI and Bharti School for providing us with the training data. We would also like to thank the International Telecommunication Union (ITU) Focus Group Autonomous Network(FG-AN) for selecting our team to present this work at the International Competition of Build Your Own Closed Loop (BYOCL).


\ifCLASSOPTIONcaptionsoff
  \newpage
\fi

\bibliographystyle{unsrt}
\bibliography{reference}

\begin{IEEEbiography}{Michael Shell}
\centering
Biography text here.
\end{IEEEbiography}

\begin{IEEEbiographynophoto}{John Doe}
\centering
Biography text here.
\end{IEEEbiographynophoto}

\begin{IEEEbiography}{First A. Author}{\space}(M'76--SM'81--F'87) and all authors may include biographies. Biographies are often not included in conference-related papers. This author became a Member (M) of IEEE in 1976, a Senior Member (SM) in 1981, and a Fellow (F) in 1987. The first paragraph may contain a place and/or date of birth (list place, then date). Next, the author’s educational background is listed. The degrees should be listed with the type of degree in what field, which institution, city, state, and country, and the year the degree was earned. The author's major field of study should be lowercase.

The second paragraph uses the pronoun of the person (he or she) and not the author's last name. It lists military and work experience, including summer and fellowship jobs. Job titles are capitalized. The current job must have a location; previous positions may be listed without one. Information concerning previous publications may be included. Try not to list more than three books or published articles. The format for listing publishers of a book within the biography is the title of the book (publisher name, year), similar to a reference. Current and previous research interests end the paragraph.

The third paragraph begins with the author's title and last name (e.g., Dr. Smith, Prof. Jones, Mr. Kajor, Ms. Hunter). List any memberships in professional societies other than the IEEE. Finally, list any awards and work for IEEE committees and publications. If a photograph is provided, it should be of good quality and professional-looking. Following are two examples of an author’s biography.
\end{IEEEbiography}

\begin{IEEEbiography}{Second B. Author}{\space}was born in Greenwich Village, New York, NY, USA in 1977. He received the B.S. and M.S. degrees in aerospace engineering from the University of Virginia, Charlottesville, in 2001 and the Ph.D. degree in mechanical engineering from Drexel University, Philadelphia, PA, in 2008.

    From 2001 to 2004, he was a Research Assistant with the Princeton Plasma Physics Laboratory. Since 2009, he has been an Assistant Professor with the Mechanical Engineering Department, at Texas A\&M University, College Station. He is the author of three books, more than 150 articles, and more than 70 inventions. His research interests include high-pressure and high-density nonthermal plasma discharge processes and applications, microscale plasma discharges, discharges in liquids, spectroscopic diagnostics, plasma propulsion, and innovation plasma applications. He is an Associate Editor of the journal {\it Earth}, {\it Moon}, {\it Planets}, and holds two patents. 

   Dr. Author was a recipient of the International Association of Geomagnetism and Aeronomy Young Scientist Award for Excellence in 2008, and the IEEE Electromagnetic Compatibility Society Best Symposium Paper Award in 2011. 
\end{IEEEbiography}

\begin{IEEEbiography}{Third C. Author, Jr.}{\space}(M’87) received a B.S. degree in mechanical engineering from National Chung Cheng University, Chiayi, Taiwan, in 2004 and an M.S. degree in mechanical engineering from National Tsing Hua University, Hsinchu, Taiwan, in 2006. He is currently pursuing a Ph.D. degree in mechanical engineering at Texas A\&M University, College Station, TX, USA.

    From 2008 to 2009, he was a Research Assistant with the Institute of Physics, Academia Sinica, Tapei, Taiwan. His research interest includes\vadjust{\vfill\pagebreak} the development of surface processing and biological/medical treatment techniques using nonthermal atmospheric pressure plasmas, fundamental study of plasma sources, and fabrication of micro- or nanostructured surfaces. 

   Mr. Author’s awards and honors include the Frew Fellowship (Australian Academy of Science), the I. I. Rabi Prize (APS), the European Frequency and Time Forum Award, the Carl Zeiss Research Award, the William F. Meggers Award, and the Adolph Lomb Medal (OSA).
\end{IEEEbiography}

\end{document}